# Etude de Modèles à base de réseaux Bayésiens pour l'aide au diagnostic de tumeurs cérébrales

Fradj Ben Lamine[1], Karim Kalti[2] et Mohamed Ali Mahjoub[3]

[1] Ecole supérieur des sciences et des technologies de Hammam Sousse - Université de Sousse,
`benlamine.fradj@gmail.com`
[2] Faculté des sciences de Monastir
Unité de recherche SAGE – Ecole Nationale d'Ingénieurs de Sousse
`karim.kalti@gmail.com`
[2] Institut Préparatoire aux Etudes d'Ingénieurs de Monastir,
Unité de recherche SAGE – Ecole Nationale d'Ingénieurs de Sousse
`medali.mahjoub@ipeim.rnu.tn`

**Résumé** : Cet article décrit différents modèles à base de réseaux bayésiens RB modélisation de l'expertise du diagnostic des tumeurs cérébrales. En effet, ces derniers sont bien adaptés à la représentation de l'incertitude qui caractérise le processus de diagnostic de ces tumeurs. Dans notre travail, nous avons d'abord testé plusieurs structures du réseau bayésien déduites du raisonnement effectué par les médecins d'une part et de structures générées automatiquement d'autre part. Cette étape a pour but de trouver la meilleure structure qui augmente la précision du diagnostic. Les algorithmes d'apprentissage automatique concernent les algorithmes MWST-EM, SEM et SEM+T. Afin d'estimer les paramètres du réseau bayésien à partir d'une base de données incomplètes, nous avons proposé une extension de l'algorithme EM en lui ajoutant des connaissances a priori sous forme des seuils calculés par la première phase de l'algorithme RBE. Les résultats obtenus très encourageants sont discutés en fin du papier.

**Mots-clés** : Réseau bayésien, Tumeurs cérébrales, Diagnostic, Apprentissage des paramètres, Apprentissage de structure, Inférence, Algorithme EM, algorithme RB.

## 1 Introduction

Une tumeur cérébrale est un groupe de cellules anormales à l'intérieur de l'encéphale. Les tumeurs primitives se forment dans le cerveau et peuvent être bénignes ou malignes. Si dans la plupart du corps les tumeurs bénignes ne représentent pas le même danger que les tumeurs malignes, dans l'encéphale les types de tumeurs peuvent être graves et mettre éventuellement la vie en danger. Ainsi, l'étude des tumeurs cérébrales possède de plus en plus une grande importance. En outre, leur diagnostic assez difficile vu le nombre de variétés de ces tumeurs. Plusieurs caractéristiques entrent en jeu pour décider le type exact de la tumeur. Certaines de ces caractéristiques sont extraites à partir des images par résonance magnétique. D'autres caractéristiques sont extraites des données cliniques du patient.

Dans ce contexte, notre travail porte sur la modélisation de l'expertise relative au diagnostic d'une famille des tumeurs cérébrales qui est celle des tumeurs de type intra-axial sus-tentorielle. Elles sont caractérisées par la forte ressemblance de leurs caractéristiques ce qui rend leur diagnostic assez difficile. Nous limitons notre champ d'études aux sept tumeurs suivantes : Astrocytome, Gliome, Oligodendrogliome, lymphome, Métastase, Ependymome, Médulloblastome. En outre, les informations qui entrent dans le diagnostic de ces tumeurs cérébrales se répartissent en deux catégories ; une concernant les données extraites du dossier médical du patient comme par exemple l'âge, le sexe et l'historique médical, et une autre constituée d'un ensemble de données extraites à partir des images IRM. L'utilisation de ces caractéristiques dans le cadre d'une consultation médicale est insuffisante. L'expertise du médecin est nécessaire afin d'avoir une bonne précision dans ce type de diagnostic. Ce dernier intervient dans la majorité de cas étudiés pour lever l'incertitude d'identification du type de la tumeur. Son intervention se base sur les situations déjà traitées.

Les modèles graphiques probabilistes semblent être un outil approprié à la modélisation des systèmes d'aide au diagnostic médical. Par nature, ces systèmes intègrent un certain degré d'incertitude et expriment de façon intuitive le phénomène de dépendance. Les réseaux bayésiens RB offrent la possibilité de rassembler et de fusionner de connaissances de diverses natures dans un même modèle (P NAïM 2007). Pour ces raisons, le formalisme des RBs a été retenu pour la modélisation de notre problème. Selon ce formalisme, les caractéristiques sont modélisées par les nœuds du réseau bayésien qui sont liés par des arcs représentant leurs dépendances. L'incertitude de l'expert est exprimée par un ensemble de tables de probabilités conditionnelles CPT qui permettent de quantifier le taux de certitude d'un nœud en tenant compte de ses parents. Vu la complexité de ce travail pour des données de grande taille, on procède aujourd'hui parapprentissage automatique à partir d'une base de données. Ce processus connu sous le nom de « apprentissage de paramètres » permet de remplir les CPTs même en présence de données manquantes.

Ce papier est organisé comme suit. Tout d'abord nous présentons un rappel très bref sur les concepts de base des réseaux bayésiens. Ensuite, nous donnons une description générale de notre modèle d'aide au diagnostic des tumeurs cérébrales. Dans la section 4, nous exposons en détail le problème du choix de la structure du réseau et les justifications nécessaires ainsi que les résultats des expérimentations. L'apprentissage des paramètres et les algorithmes associés sont traités dans la section 5. Enfin nous terminons par une conclusion.

## 2   Concepts de base sur les réseaux bayésiens

Le principe associé aux réseaux bayésiens est de tenir compte des indépendances conditionnelles pour simplifier la loi jointe découlant du théorème de Bayes généralisé. Un réseau bayésien permet la description qualitative et quantitative des indépendances conditionnelles entre variables.
- Description qualitative: un réseau bayésien est un graphe acyclique dirigé $G = (V, E)$ où V représente l'ensemble des variables aléatoires $(V_1, V_2, \cdots, V_n)$ décrivant les évènements du domaine, et E est l'ensemble des arcs. Un arc de $V_i$ vers $V_j$ constitue une relation de dépendance entre ces nœuds.
- Description quantitative : il s'agit d'un ensemble des tables de probabilités conditionnelles associées à chaque variable $V_i$. Ces tables sont fournies soit par un expert du domaine soit par un apprentissage automatique.

D'une façon formelle, un réseau bayésien est défini par :
- un graphe acyclique orienté (DAG) $G = (V, E)$, où V est l'ensemble des nœuds de G, et E l'ensemble des arcs de G ;
- un espace probabilisé fini $(\Omega, Z, p)$ ;
- un ensemble de variables aléatoires associées aux nœuds du graphe et défini sur $(\Omega, Z, p)$, tel que :

$$p(V_1, V_2, \cdots, V_n) = \prod_{i=1}^{n} p(V_i / C(V_i))$$

où $C(V_i)$ est l'ensemble des parents de $V_i$ dans le graphe G.

Lors de la construction de réseaux bayésiens, il n'est pas toujours évident qu'un expert puisse fournir de façon numérique l'ensemble des paramètres nécessaires à l'inférence dans un graphe. C'est grâce à la méthode d'apprentissage qu'on remédie ce problème.

## 3   Présentation du modèle proposé

Les étapes de base nécessaires à la proposition d'un modèle d'aide au diagnostic des tumeurs cérébrales en utilisant les réseaux bayésiens sont :

la modélisation de la structure, la modélisation des paramètres par apprentissage et l'inférence.

La modélisation de la structure du réseau bayésien est une phase assez complexe. Cette difficulté est due au grand nombre de caractéristiques qui entrent en jeu dans le diagnostic des tumeurs cérébrales. En outre, nous procédons à l'estimation des paramètres par apprentissage automatique. Pour ce faire, nous avons collecté un ensemble de cas traités auparavant qui sont sous la forme de rapports médicaux et des images IRM (77 cas). La troisième phase est une étape qui permet de valider la structure et les paramètres du réseau bayésien. Dans notre travail, nous n'allons pas étudier cette étape pour la grande variété de ces algorithmes. Nous avons choisi un algorithme standard appelé « arbre de jonction » pour faire l'inférence. Le calcul exact effectué par cet algorithme est derrière ce choix.

Chaque étape citée ci-dessus peut influencer la qualité du modèle. Nos travaux de recherche mettent plutôt l'accent sur les deux premières étapes. Le développement de notre approche se déroule en deux phases (figure 1). Dans la première phase, nous nous focalisons sur la modélisation de la structure du réseau bayésien. Pour ce faire, nous partons d'une structure déduite des entretiens que nous avons effectués avec les médecins pour identifier les dépendances entre les caractéristiques. Nous améliorons ensuite cette structure de départ en lui apportant des modifications basées sur un ensemble de descripteurs prioritaires d'identification des tumeurs cérébrales.

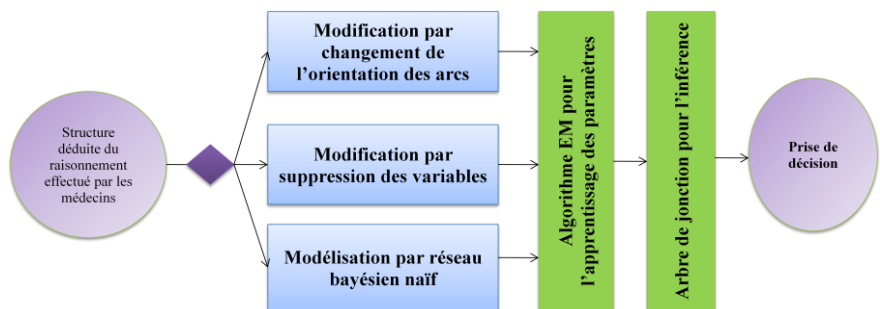

FIGURE 1–*Démarche de la première étape*

Dans la deuxième phase, nous nous intéressons essentiellement à la modélisation des paramètres du réseau. Pour ce faire, nous fixons d'abord la structure de ce dernier. Puis, nous déduisons ses paramètres à travers une technique d'apprentissage que nous proposons. Cette technique est une variante de l'algorithme EM (DEMPSTER 1977, RM NEAL 1998, MA MAHJOUB 2011).

## 4  Modélisation de la structure du réseau bayésien

La modélisation de la structure du réseau bayésien possède une grande importance de point de vue conceptuel. Elle permet de concevoir les connaissances utilisées sur cette problématique de manière simple et compréhensible par les non spécialistes. Pour ce faire, nous sommes partis d'une proposition de structure directement déduite à partir du raisonnement effectué par les médecins comprenant trente nœuds répartis sur quatre niveaux. Le premier niveau représente l'ensemble des caractéristiques qui sont mises en jeu pour la prise de décision dans le cadre de diagnostic des tumeurs cérébrales. Le deuxième et le troisième niveau sont des nœuds intermédiaires de décision. Le quatrième niveau est composé d'un seul nœud qui désigne le nœud décision permettant d'identifier le type de la tumeur cérébrale. La structure de ce réseau est donnée par la figure 2.

La librairie Bayesian Network Toolbox (BNT) de Kevin Murphy. Utilisée dans les expérimentations permet de donner la courbe représentative de la variation de Log-Likelihood qui est définie par :

$$\text{LL}(D|\theta) = \log L(D|\theta) = \sum_{i=1}^{n}\sum_{j=1}^{q_i}\sum_{k=1}^{r_i} N_{i,j,k} \log \theta_{i,j,k} \quad (1)$$

Où : $n$ est le nombre de nœuds, $q_i$ est le nombre de configurations de parents du nœud i, $r_i$ est le nombre d'états du nœud i, $N_{i,j,k}$ est le nombre de cas où le nœud $i$ est dans l'état $k$ et ses parents dans la configuration $j$.

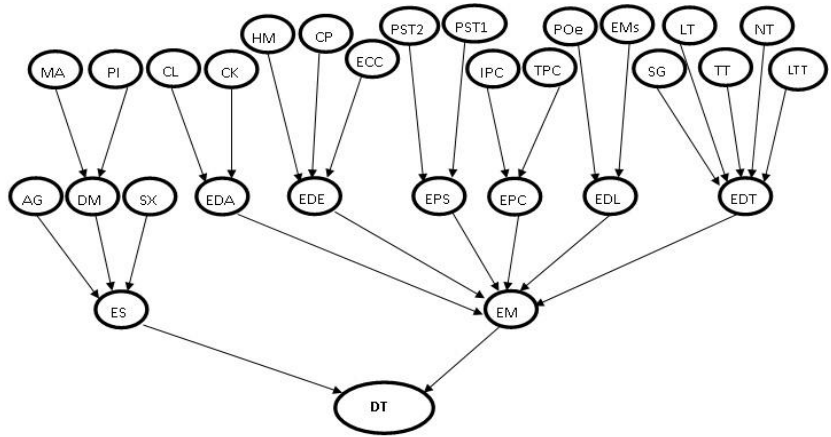

FIGURE 2–*Structure du réseau bayésien déduite à partir du raisonnement effectué par les médecins*

TABLE 1– *Signification des nœuds*

| Nom du nœud | Signification | Nom du nœud | Signification |
|---|---|---|---|
| AG | Age | ES | Etat Clinique |
| CK | Composante Kystique | HM | Hémorragie |
| CL | Calcification | IPC | Importance Prise de Contraste |
| CP | Composition | LT | Location Tumeur |
| DM | Dossier Médical | LTT | Limite Tumeur |
| DT | Décision Tumeur | MA | Maladies auxiliaires |
| ECC | Envahissement Corps Calleux | NT | Nombre Tumeur |
| EDA | Etat données Auxiliaires | PI | Première Infection |
| EDE | Etat données encéphaliques | Poe | Présence Œdème |
| EDL | Etat données Liquides | PST1 | Prise de Signal en T1 |
| EDT | Etat Données tumeur | PST2 | Prise de Signal en T2 |
| EM | Etat Radiologique | SG | Siège |
| Ems | Effet de Masse | SX | Sexe |
| EPC | Etat de Prise de Contraste | TPC | Type de Prise de Contraste |
| EPS | Etat prise de signal | TT | Taille Tumeur |

Une table de probabilité conditionnelle est un vecteur multidimensionnel noté $\Theta$. Cette table contient l'ensemble de probabilités de la variable $X_i$ pour chacune de ses valeurs possibles sachant chacune des valeurs prises par l'ensemble de ses parents $pa(X_i)$. Chaque composante du vecteur $\Theta$ est exprimée par :

$$\theta_{i,j,k} = P\left(X_i = x_k \mid pa\left(X_i = x_j\right)\right)$$

$\theta_{i,j,k}$ est la valeur du paramètre où le nœud *i* est dans l'état *k* et ses parents sont dans la configuration *j*. L'équation 1 permet de donner la performance de paramètres calculés dans le réseau bayésien. Le Log-Likelihood donne les paramètres qui décrivent le mieux la base d'apprentissage. Cette valeur est mise à jour à chaque itération dans l'algorithme EM. Notons que dans la suite de ce paragraphe, nous adoptons l'algorithme EM pour faire l'apprentissage de paramètres et l'algorithme d'arbre de jonction pour l'inférence.

### 4.1 Etude Expérimentale sur le choix de la structure

Nous commençons par évaluer la performance de la structure déduite à partir du raisonnement effectué par les médecins. Nous utilisons la base d'apprentissage (60 cas) et une base de test de 17 instanciations possibles pour les nœuds. Des indicateurs ont été calculés et ce relatif à la structure du réseau bayésien déduite à partir du raisonnement effectué par les médecins concernant le nombre d'itérations (32) de l'algorithme EM, le temps d'exécution (64.107) et la précision[1] de prédiction de l'inférence (40%). Ces résultats obtenus pour cette première proposition de la structure ne sont pas satisfaisants de point de vue précision d'inférence. Ceci peut être expliqué par le nombre important de données manquantes au niveau des nœuds intermédiaires. Pour faire face au faible nombre de cas qui existent dans la base de données, nous avons ajouté des connaissances a priori sous la forme des coefficients de Dirichlet (E. BAUER 1997, M. RAMONI 1997) qui consiste à créer de cas imaginaires avec le même nombre de variables et les mêmes valeurs prises par ces nœuds. L'ajout des connaissances a priori de Dirichlet ne fait pas varier le Log-Likelihood.

TABLE 2–*Effectif des tumeurs en pourcentage*

| Type de la tumeur | Effectif (%) |
|---|---|
| Tumeur 1 | 16.66 |
| Tumeur 2 | 11.11 |
| Tumeur 3 | 31.94 |
| Tumeur 4 | 9.72 |
| Tumeur 5 | 2.77 |
| Tumeur 6 | 22.22 |
| Tumeur 7 | 0 |
| Tumeur 8 (Autres Tumeur) | 5.58 |

Les valeurs des indicateurs sont obtenues après la phase d'apprentissage et d'inférence en ajoutant des connaissances a priori de Dirichlet. Ces valeurs se rapprochent de celles obtenues à partir de la structure initiale. Les coefficients de Dirichlet sont inspirés de la base d'apprentissage ce qui explique les valeurs obtenues. Cependant, la prise de décision avec ces modélisations reste ambigüe.
Ceci nous donne des motivations pour trouver de nouvelles structures qui augmentent la précision dans le diagnostic des tumeurs cérébrales. Ces modifications opèrent par :
- ✓ Changement de l'orientation des arcs en conservant le même nombre de variables
- ✓ Elimination des variables
- ✓ Modélisation par réseau bayésien naïf
- ✓ Modélisation par apprentissage automatique de structures de réseaux bayésiens

---

[1] La précision de l'inférence est le pourcentage d'avoir un résultat correct du diagnostic des tumeurs cérébrales lorsque on injecte certaines valeurs dans le réseau bayésien.

Chaque modification possède une stratégie particulière. Les experts de diagnostic des tumeurs cérébrales ont donné un tableau récapitulatif des caractéristiques prioritaires pour l'identification de six tumeurs parmi sept.

La définition des stratégies appliquées à la modification de la structure déduite à partir du raisonnement effectué par les médecins est basée sur les connaissances collectées. Dans le paragraphe suivant, nous allons décrire le principe de chaque stratégie ainsi que les résultats obtenus.

### 4.1.1 Proposition de nouvelles structures par changement de l'orientation des arcs

Le principe consiste à dégager les caractéristiques prioritaires dans le diagnostic des tumeurs cérébrales et de les orienter directement au nœud décision. On signale à cet effet, que plusieurs critères entrent dans l'aide à la décision des tumeurs cérébrales. On cite dans un premier plan le siège, l'œdème et la calcification. Nous proposons dans la suite des modélisations possibles dérivées de la structure déduite à partir du raisonnement effectué par les médecins en changeant l'orientation des arcs qui sont en relation avec ces nœuds.

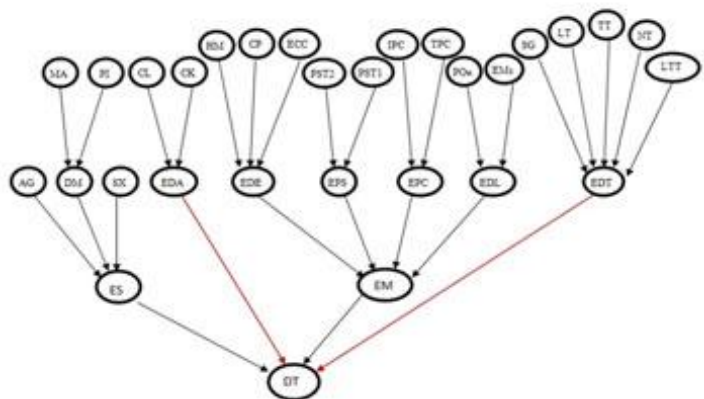

FIGURE 3–*Proposition d'une structure en changeant les nœuds «eda » et « edt »
Nombre d'itérations 21 ; temps d'exécution 41.13 s Précision d'inférence 40%*

La figure 3 illustre le changement de la direction d'arc de deux nœuds « eda » et « edt » vers le nœud décision au lieu d'être dirigée vers le nœud « em ». Ces nœuds rassemblent la majorité des caractéristiques qui entrent dans le diagnostic des tumeurs cérébrales de façon directe. Nous remarquons d'après les premières expérimentations que le nombre d'itérations pour l'algorithme EM diminue par rapport à la structure initiale. Ceci est du au nombre de paramètres qui a diminué dans le réseau bayésien. Le résultat de la précision dans l'inférence reste à 40%.

En outre, les nœuds « sg », « lt » et « cl » sont les trois caractéristiques similaires dans les tumeurs étudiées. En effet, ces nœuds jouent un rôle important dans la prise de décision. L'étude de la performance de la structure associée aux changements des arcs nous montre que le pourcentage de précision dans l'inférence est similaire à celui des modélisations étudiées auparavant. Le seul avantage de cette suggestion est le gain en temps d'exécution. D'autre part, après expérimentations, les différentes modifications de la structure initiale proposée par les médecins montrent que l'amélioration dans le pourcentage de précision effectué par l'inférence reste très faible et n'obéit pas aux tendances du modèle d'aide au diagnostic. Ainsi nous avons procédé par modification basée sur la suppression de certaines variables.

### 4.1.2 Proposition de nouvelles structures par suppression de variables

Le principe consiste à éliminer les variables qui ne sont pas prioritaires dans le diagnostic ou celles qui possèdent beaucoup des valeurs manquantes. Nous avons proposé plusieurs structures dans ce choix. Retenons les meilleures suggestions. En supprimant les nœuds « ma » et « pi » la structure obtenue est la même que la structure initiale mais en éliminant les nœuds « ma » et « pi » qui sont les maladies auxiliaires et la première infection. Ces deux caractéristiques n'entrent pas dans le diagnostic de la majorité des tumeurs. En outre, elles présentent des cas qui ne sont pas observés dans la base d'apprentissage, ce qui explique leur élimination de la structure des médecins. Nous constatons que le Log-Likelihood dépasse la valeur -1000. Ceci s'explique par la réduction des valeurs manquantes dans la base d'apprentissage. Le faible pourcentage de précision obtenu (40%) de l'inférence nous encourage à supprimer d'autres variables afin d'augmenter ce taux. Après suppression aussi des nœuds « edt » et « eda », on remarque que les résultats obtenus sont très encourageants de point de vue de l'inférence (80%). De plus, l'algorithme EM converge dans un nombre d'itérations égal à cinq. Le temps d'exécution est évalué à 9,90 secondes. Les résultats obtenus nous montrent la variation de l'allure pour le Log-Likelihood qui atteint la valeur -900. Les structures proposées précédemment, nous conduisent à modifier la structure du réseau bayésien de façon définitive. Nous passons ainsi aux réseaux bayésien de classification ou réseaux naïfs.

### 4.1.3 Modélisation à l'aide d'un réseau bayésien naïf

Rappelons tout d'abord que le réseau bayésien naïf est directement issu de l'application de la règle de décision de Bayes en rajoutant l'hypothèse d'indépendance conditionnelle des caractéristiques de la tumeur cérébrale conditionnellement au caractéristiques du diagnostic. Un des inconvénients de ces réseaux est le grand nombre de paramètres à estimer, alors que dans la plupart des cas, le nombre de données disponibles est faible. La structure du réseau se décompose en deux couches dont la première est constituée par un seul nœud qui présente les différentes classes de décision et la deuxième regroupe les différentes caractéristiques du problème. Les résultats obtenus sont meilleurs que les précédents en améliorant la précision de l'inférence (88% des diagnostics corrects). Ceci est dû au faible nombre des données manquantes dans la base d'apprentissage. Ces résultats d'inférence montrent la limite de la structure déduite à partir du raisonnement effectué par les médecins et mettent par conséquent en cause l'utilité des nœuds de décision intermédiaires.

En outre, l'hypothèse d'indépendance entre les attributs utilisée dans le classifier de Bayes naïf est généralement fausse. Il existe différentes techniques pour assouplir cette hypothèse. Elles consistent à identifier les dépendances conditionnelles entre les attributs. Nous obtenons alors une sous-structure optimale sur les observations en adaptant la méthode de recherche de l'arbre de recouvrement de poids maximal. Cette méthode s'applique à la recherche de structure d'un réseau bayésien en fixant un poids à chaque arête potentielle de l'arbre. Une fois cette matrice de poids définie, il suffit d'utiliser un des algorithmes standards de résolution du problème de l'arbre de poids maximal. L'arbre non dirigé retourné par cet algorithme doit ensuite être dirigé en choisissant une racine puis en parcourant l'arbre par une recherche en profondeur. La racine peut être choisie soit aléatoirement, soit à l'aide de connaissance a priori. Nous obtenons le TAN (Tree Augmented Naive Bayes) (figure4). D'après nos expérimentations, on a constaté que le réseau bayésien naïf a donné un meilleur résultat que le TAN (12 diagnostics corrects sur 17 soit 70%). En effet, d'une part la sélection de l'attribut racine est importante pour construire un TAN. D'autre part, des liens non nécessaires peuvent exister dans un TAN. Ainsi nous pouvons définir l'algorithme TAN comme suit :

1- Nous choisissons l'attribut $A_{racine}$ qui a l'information mutuelle maximale avec la classe,

2- Nous filtrons les liens qui ont des informations mutuelles conditionnelles inférieur à un seuil. À notre compréhension, ces liens ont un risque élevé pour sur adapter les données et ensuite l'évaluation de probabilité. Puisque la structure du modèle résultant n'est pas un arbre strict, l'algorithme modifié est appelé naïve bayes augmenté par une forêt

FAN (Forest Augmented Naive bayes). Par application du FAN, on est arrivé à avoir 15 diagnostics corrects (88%). Néanmoins, pour une meilleure évaluation des modèles NB, TAN et FAN (K. JAYECH 2010), nous devons disposer d'un nombre assez élevé de dossiers de patients.

### 4.1.4 Modélisation à l'aide de réseaux bayésiens par apprentissage de structures

Nous avons testé les algorithmes suivants (O. FRANÇOIS 2006) :
– MWST-EM avec une racine aléatoire et le score QBIC. Il s'agit de la recherche de la structure arborescente optimale reliant toutes les variables+l'utilisation de l'algorithme EM.
– SEM initialisé avec une chaîne, score QBIC. C'est une adaptation de l'algorithme EM pour l'apprentissage de la structure (Structural EM). La méthode SEM est une technique gloutonne (donc itérative) qui utilise un algorithme EM (donc itératif) à l'intérieur de chacune de ses itérations. Elle requiert donc un temps de calcul important
– SEM+T initialisé par une structure arborescente optimale, score QBIC. C'est une initialisation de l'algorithme glouton SEM par l'arbre obtenu par la méthode MWST-EM. La méthode résultante est nommée SEM+T.

Nous avons dressé dans la figure 5 la variation de la fonction de log de vraisemblance pour les différents algorithmes comparés à la structure du médecin. D'autre part, nous remarquons que les meilleures précisions correspondant aux structures de FAN et un peu moins MWST-EM (figure 6).

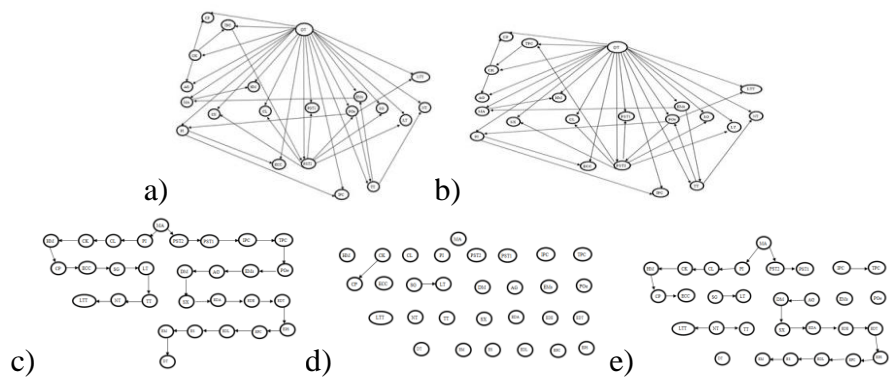

FIGURE 4–*Structures de réseaux apprises par notre modèle*
a) TAN b) FAN c) MSWT-EM d) SEM e) SEM+T

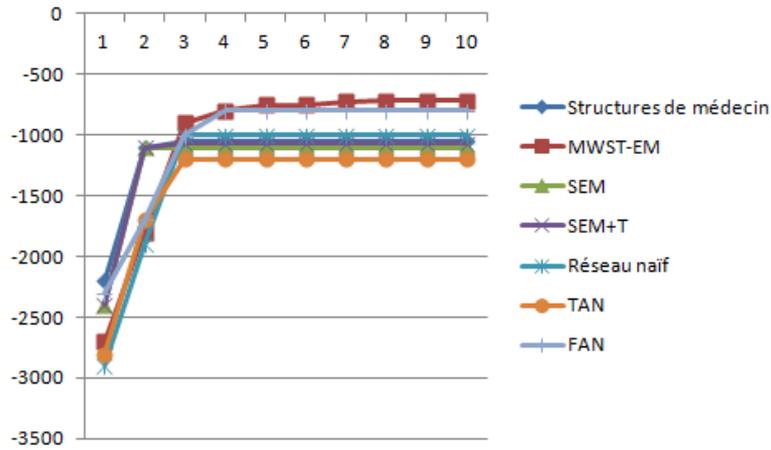

FIGURE 5–*Variation de Log-Likelihood pour l'apprentissage automatique de la structure*

## 5 Apprentissage des paramètres du réseau bayésien

Les sources d'incertitude dans le cadre d'un diagnostic médical sont présentes dans la majorité de modèles d'intelligence artificielle. L'incertitude peut être provoquée par l'absence d'un historique clinique complet et détaillé pour les patients. En plus, ces derniers ne se rappellent pas parfois de tous les symptômes pendant l'évaluation de leurs maladies.

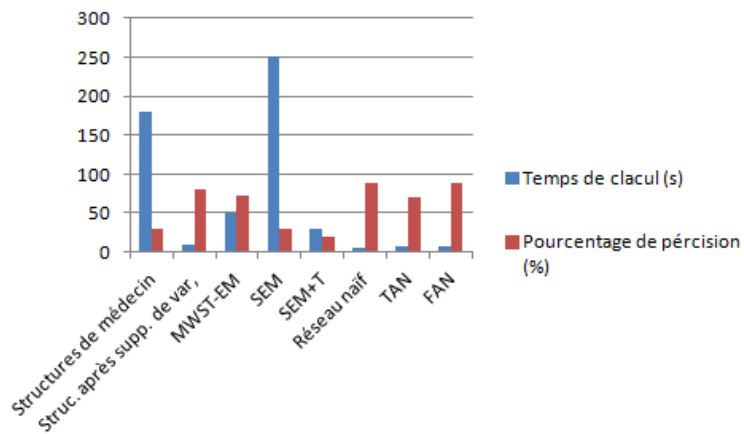

FIGURE 6–*Variation du temps de calcul et de la précision*

Pendant la phase d'interrogation, l'information donnée par le patient peut être mal exprimée et il peut même mentir au médecin. En médecine, beaucoup de données sont imprécises en particulier celles qui proviennent des images IRM et de tests de laboratoire. Ces causes conduisent à une base d'apprentissage souvent incomplète qui nécessite des algorithmes appropriés pour travailler dessus. Ce genre de problème est traité dans (RS. NICULESCU & AL 2006, M. RAMONI 1997).Dans ce qui suit, nous donnons une contribution qui entre dans ce cadre. En effet, nous proposons un algorithme d'apprentissage pouvant travailler sur des données manquantes. Cet algorithme est une extension de l'algorithme standard d'apprentissage EM en régularisant les paramètres maximisés par des seuils extraits de l'exécution de la première phase de l'algorithme RBE. Rappelons ici que l'approche Robust Bayesian Estimator RBE est considérée comme une procédure qui parcourt tout l'ensemble des données D, enregistre les observations concernant les variables et ensuite, elle permet de borner la probabilité conditionnelle d'une variable $X_i$. On pourra consulter par exemple (AD. FEELDERS 2006) pour plus de détails. Notre approche est une combinaison de l'algorithme EM et de la première phase de l'algorithme RBE. Son objectif est de corriger la valeur de chaque paramètre $\theta_{i,j,k}$ après la phase de maximisation de l'algorithme EM. Admettons que nous avons : $\min_{i,j,k}$ = la probabilité minimale calculée suivant la phase 1 de l'algorithme RBE et $\max_{i,j,k}$= la probabilité maximale calculée suivant la phase 1 de l'algorithme RBE.

Le principe général consiste à augmenter la valeur $\theta_{i,j,k}$ si elle est inférieure à $\min_{i,j,k}$ avec cette valeur. De même, si la valeur de $\theta_{i,j,k}$ dépasse la probabilité maximale calculée par la phase 1 de l'algorithme RBE, nous la mettons à jour avec $\max_{i,j,k}$. Le but majeur de cette approche est de régulariser les valeurs de $\theta_{i,j,k}$ présentes dans le réseau bayésien afin de minimiser le taux d'erreur entre la valeur de $\theta_{i,j,k}$ maximisée et la contrainte des bornes utilisés par de l'algorithme RBE : $\min_{i,j,k} \leq \theta_{i,j,k} \leq \max_{i,j,k}$. Afin de respecter la contrainte que la somme des probabilités, pour une combinaison des parents « j » pour toutes les valeurs possibles d'état du nœud « i », soit égale à 1, nous avons fait une normalisation des nouvelles valeurs calculées selon la formule suivante :

$$\theta_{i,j,k}^{t+1} = \frac{\theta_{i,j,k}^t}{\sum_i \theta_{i,j,k}^t} \qquad (2)$$

Où : $\theta_{i,j,k}^{(t)}$ : le paramètre où la variable $X_i$ est dans l'état $x_k$ et ses parents sont dans la configuration $x_j$ à l'itération t. *k'* : est l'état du nœud $X_i$.

L'algorithme EMS proposé (Espérance Maximisation Seuillage) décompose en quatre étapes ; i) Espérance, similaire à l'algorithme EM, ii) Maximisation, similaire aussi à l'algorithme EM, iii) Seuillage et iv) Normalisation selon l'équation (2).

## 5.2 Résultats de l'apprentissage des paramètres

Nous avons comparé notre contribution avec le standard d'apprentissage de paramètres EM. Nous constatons que l'allure de la courbe de l'algorithme EM est supérieure à l'algorithme EMS de point de vue Log-Likelihood. Ceci est dû à la phase de seuillage qui augmente certaines valeurs des paramètres. D'où le Log-Likelihood augmente légèrement par rapport à celui de l'algorithme EM. La figure 7 illustre cette comparaison. D'autre part, nous remarquons que l'algorithme EMS converge en un nombre d'itérations inférieur à celui de l'algorithme EM pour le même point de départ. Cette rapidité de convergence est assurée dans 70% de cas si nous adoptons plusieurs points de départ pour les deux algorithmes. Le temps d'exécution pour l'algorithme EMS est nécessairement plus grand que celui de l'algorithme EM. Pour l'inférence, les deux algorithmes sont presque semblables (figure 8).

Le but majeur de l'ajout de cette phase de seuillage est d'avoir des paramètres qui vérifient l'appartenance à l'intervalle calculé par la première phase de l'algorithme RBE et en utilisant l'algorithme EM comme algorithme d'apprentissage. Nous montrons dans la figure 9 la distribution de probabilité dans le nœud décision. Ce nœud contient huit valeurs représentant huit types de tumeurs. Il possède deux parents dont chacun contient deux valeurs. Ceci conduit à quatre combinaisons pour chaque type de tumeur. L'algorithme EMS augmente le pourcentage de vérification des contraintes extraites par la première phase de l'algorithme RBE en le comparant avec l'algorithme EM mais n'atteint jamais le pourcentage maximal. Si cette contrainte n'est pas vérifiée, nous constatons que la valeur du paramètre essaye de se rapprocher de l'intervalle calculé.

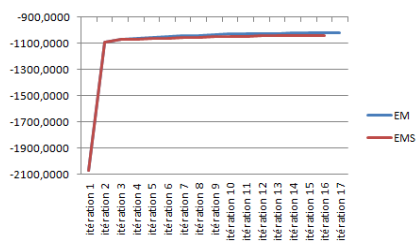
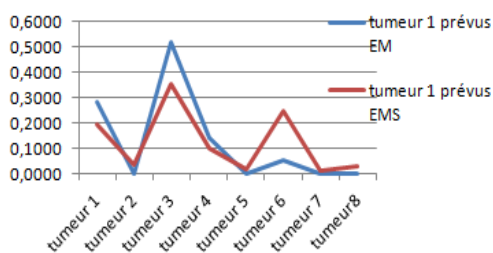

FIGURE 7–*Variation Log-Likelihood de EM et EMS*

FIGURE 8–*Variation des probabilités après inférence en utilisant EM et EMS par rapport aux tumeurs prévues*

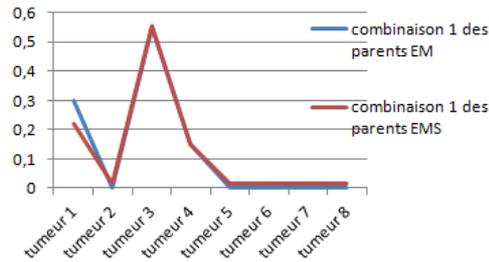

FIGURE 9–Variation de la distribution des probabilités dans le nœud décision avec EM et EMS

## 6   Conclusion

Dans le cadre de ce travail, nous avons proposé une modélisation de l'expertise du diagnostic des tumeurs cérébrales à l'aide de réseaux bayésiens. En premier lieu, nous sommes partis d'une structure déduite à partir du raisonnement effectué par les médecins. Nous avons effectué des modifications entre les dépendances des caractéristiques pour améliorer la précision du diagnostic. Ces changements nous ont conduits à utiliser des dépendances directes entre les caractéristiques pour le diagnostic et le nœud décision. Cette modélisation est effectuée dans un premier temps par le réseau bayésien naïf avec trois variantes (NB, TAN, FAN) dont les résultats ont donné la meilleure précision. Dans un deuxième temps, nous avons utilisé des algorithmes d'apprentissage de structures à savoir MWST-EM, SEM, et SEM+T. Nous remarquons que les meilleures précisions correspondent aux structures de NB, FAN et un peu moins MWST-EM. Le faible taux enregistré par les méthodes d'apprentissage automatique de structures est sans doute du à la taille relativement petite de la base d'apprentissage.

Le problème d'estimation des paramètres dans le réseau bayésien nécessite une structure fixe et une base de tests. Dans notre cas, cette base présente des données manquantes. Nous avons constaté que les algorithmes qui traitent cette problématique possèdent des limites à cet égard, en particulier lorsque le nombre des données manquantes est énorme et la taille du réseau est grande. Pour remédier à cela, nous avons proposé une nouvelle approche pour l'estimation des paramètres. Elle consiste en l'extension de l'algorithme EM en lui ajoutant une troisième phase pour la correction des paramètres après l'exécution de cet algorithme. La troisième phase consiste à borner les paramètres par des seuils calculés par la première phase de l'algorithme RBE.

Nous avons comparé cette extension avec l'algorithme EM. Les résultats ont montré que notre proposition converge plus rapidement avec un pourcentage plus important que l'algorithme EM. En plus, l'étude des tables des probabilités conditionnelles montre l'absence des probabilités nulles et la vérification des bornes calculées par la première phase de l'algorithme RBE dans la majorité des paramètres du réseau bayésien.

**Références**